\documentclass{article}
\usepackage{spconf,amsmath,graphicx}

\usepackage{array}
\usepackage{amssymb}
\usepackage{amsmath}
\usepackage{graphicx}
\usepackage{float}
\usepackage{subfigure}
\usepackage{color}
\usepackage{amsthm,bm,amsfonts}
\usepackage{booktabs}
\usepackage[linesnumbered,ruled]{algorithm2e}
\usepackage{diagbox}
\usepackage{newfloat}
\usepackage{listings}

\title{PA-GM: Position-Aware Learning of Embedding Networks for Deep Graph Matching}
\name{Dongdong Chen$^{1}$, Yuxing Dai$^{2}$, Lichi Zhang$^{1}$, Zhihong Zhang$^{2}$, Edwin R. Hancock$^{3}$}
  
\address{$^{1}$ Shanghai Jiao Tong University \\
      $^{2}$ Xiamen University \\
      $^{3}$ University of York}

\begin{document}
%
\maketitle
\begin{abstract}
Graph matching can be formalized as a combinatorial optimization problem, where there are corresponding relationships between pairs of nodes that can be represented as edges.
This problem becomes challenging when there are potential ambiguities present due to nodes and edges with high similarity, and there is a need to find accurate results for similar content matching.
In this paper, we introduce a novel end-to-end neural network that can map the linear assignment problem into a high-dimensional space augmented with node-level relative position information, which is crucial for improving the method’s performance for similar content matching.
Our model constructs the anchor set for the relative position of nodes and then aggregates the feature information of the target node and each anchor node based on a  measure of relative position. It then learns the node feature representation by integrating the topological structure and the relative position information, thus realizing the linear assignment between the two graphs.
To verify the effectiveness and generalizability of our method, we conduct graph matching experiments, including cross-category matching, on different real-world datasets. Comparisons with different baselines demonstrate the superiority of our method. Our source code is available under https://github.com/anonymous.
\end{abstract}
\section{Introduction}
\label{sec:intro}

Graph matching aims to establish the relationship between two or more graphs based on information derived from their nodes and edges \cite{gold1996graduated}.
Due to its ability to better express and encode data relationships, graph matching has gained considerable traction in computer vision and related fields.
Applications of graph matching techniques  include image registration in medical image analysis \cite{deng2010retinal}, link analysis  in social networks \cite{zhang2015multiple} and image extrapolation in computer vision\cite{wang2014biggerpicture}.
Current methods for solving the graph matching problem can be divided into two main classes: a)  learning-free or b) learning-based \cite{pachauri2013solving,yan2020learning}. 
In the first case, the critical element is the mathematical method used for obtaining an approximate solution to an intrinsically  NP-hard problem.
On the other hand, learning-based methods aim to improve the solver with a number of methods including deep neural networks.

\begin{figure}[t]
	\begin{center}
		{\centering\includegraphics[width=0.9\columnwidth]{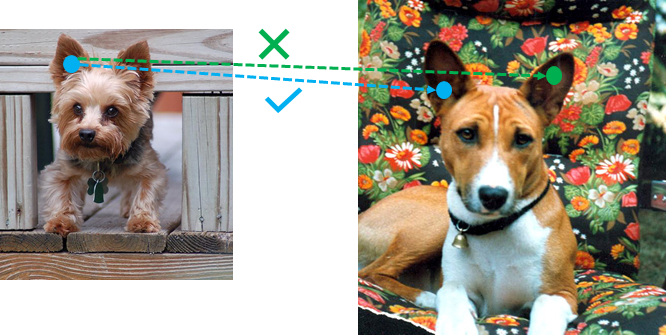}}
		\caption{A failed example of mismatching the position of the animal dog’s ears in the two images.}
	    \label{fig1}
	\end{center}
	\vspace{-2.0em}
\end{figure}

However, these methods can only compute a similarity score for the whole graph, or rely on inefficient global matching procedures\cite{xu2019cross}. For example, \cite{WangICCV19} only considers the embedding of local information for nodes in the graph, leading to a tendency to inconsistently match similar nodes from different regions of the graph, and thus resulted to have ambiguities. As is illustrated in Fig. \ref{fig1}, it shows that the node embedding representation, which relies only on local structural information and semantic node information, lacks sufficient discrimination to effectively resolve these ambiguities. As a result, it is difficult to distinguish the left and right ears of the dog in the example. From this, relative positional information is the key to the matching of diagrams, especially in cases where the graphs have similar semantic and structural content.

In this paper, we introduce the idea of position-awareness to solve the above-mentioned problem in graph matching. 
We first construct the graph as input to the model with node features extracted from an image,  and then extract a collection of anchor points as reference coordinates for each node in the graph. We further propose a  corresponding position-aware node embedding algorithm to capture the relative positional information of the nodes in the graph.
With the  node-wise  graph embedding to hand, we identify a node permutation for node-to-node correspondence.

\begin{figure*}[t]
\begin{center}
\includegraphics[width=0.85\textwidth]{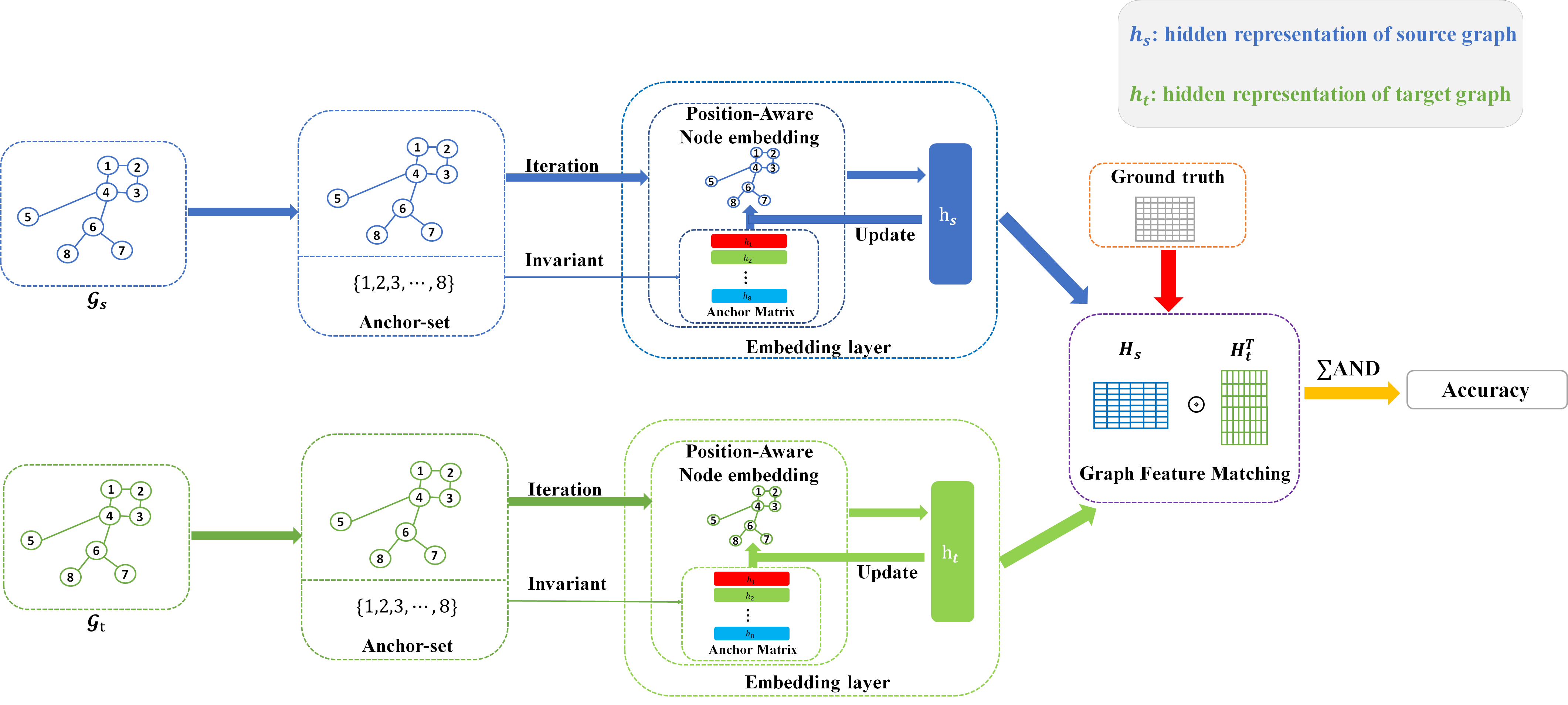}
\end{center}
\caption{Overview of the end-to-end position embedding networks for deep graph matching. The blue source graph $\mathcal{G}_s$ and the green target graph $\mathcal{G}_t$ are extracted with the node-wise graph feature representation in high-level through the two frameworks of position-aware node embedding.}
\label{figure2}
\vspace{-1.0em}
\end{figure*}

The main contributions of this paper are as follows:

\begin{itemize}
\item We propose a novel Position-Aware Learning of Embedding Network for Deep Graph Matching (PA-GM). To the best of our knowledge, there are no methods that consider the learning of an embedding augmented with relative positional information in graph matching tasks.
This restriction hampers their applications in terms of matching accuracy.

\item We extract the relative positional information for the nodes in constructing the node embedding instead of the traditional neighborhood aggregation approach. We fully consider the positional information relevant to the matching of keypoints, which is proved to be effective in our experiments.

\item The proposed framework is both scalable and flexible, and benefits from the use of a  relative position coefficient together with an alignment loss. Experiments show that the model achieves state-of-the-art results on real-world datasets.

\end{itemize}

\section{Methods}
\label{sec:methods}

In this paper, we intend to resolve the graph matching problem based on the supervised matching of graphs. Specifically, we aim to learn an end-to-end model which can extract graph information and their matches through given pair-wise ground-truth correspondences for a set of graphs, and which can be further generalized to unseen graph pairs.
The overall pipeline for graph matching using the position-aware embedding network is presented in Fig. \ref{figure2}.

\subsection{Problem Definition}
In this paper, we denote a graph by the triple $\mathcal{G}=(V, A, X)$ which consists of a finite set of nodes $V$, an adjacency matrix $A$, and a set of node attributes $X$ extracted from images using  CNN-based models\cite{simonyan2014very,qassim2018compressed}. 
We construct a vector $\mathbf{v} \in \{0,1\}^{nm \times 1}$ to indicate the match of vertices in the source graph $\mathcal{G}_s=(V_s, A_s, X_s)$ and the target graph $\mathcal{G}_t=(V_t, A_t, X_t)$. The vector has elements $\mathbf{v}_{i,j}=1$ if vertex $i \in V_s$ is matched to vertex $j \in V_t$, and $\mathbf{v}_{i,j}=0$ if otherwise. It is worth noting that all the vertex matches are subject to one-to-one mapping constraints $\sum_{j \in V_{t}} \mathbf{v}_{i, j}=1 \;  \forall i \in V_{s}$ and $\sum_{i \in V_{s}} \mathbf{v}_{i, j} \leq 1 \;  \forall j \in V_{t}$. Furthermore, we construct a square symmetric positive matrix $M \in \mathbb{R}^{nm \times nm}$ as the  affinity matrix, to encode the edge-to-edge affinity between two graphs in the  off-diagonal elements. In this way, the two-graph matching between $\mathcal{G}_s$ and $\mathcal{G}_t$ can be formulated as an edge-preserving, quadratic assignment programming (QAP) problem \cite{zhou2012factorized}:

\begin{equation}\label{eq1}
\begin{aligned}
&\underset{\mathbf{v}}{\operatorname{argmax}} \ \mathbf{v}^{\top} M \mathbf{v} \\
&\text { s.t. } \sum_{j \in V_{t}} \mathbf{v}_{i, j}=1 \;  \forall i \in V_{s}, \sum_{i \in V_{s}} \mathbf{v}_{i, j} \leq 1 \;  \forall j \in V_{t},
\end{aligned}
\end{equation}
where $\mathbf{v} \in \{0,1\}^{nm \times 1}$ and $M \in \mathbb{R}^{nm \times nm}$. 
The goal of graph matching is to establish a correspondence between two attributed graphs, which minimizes the sum of local and geometric costs of assignment between the vertices of the two graphs.

\subsection{Position-Aware Node Embedding}
\begin{figure}[t]
		\centering
		\includegraphics[width=1\columnwidth]{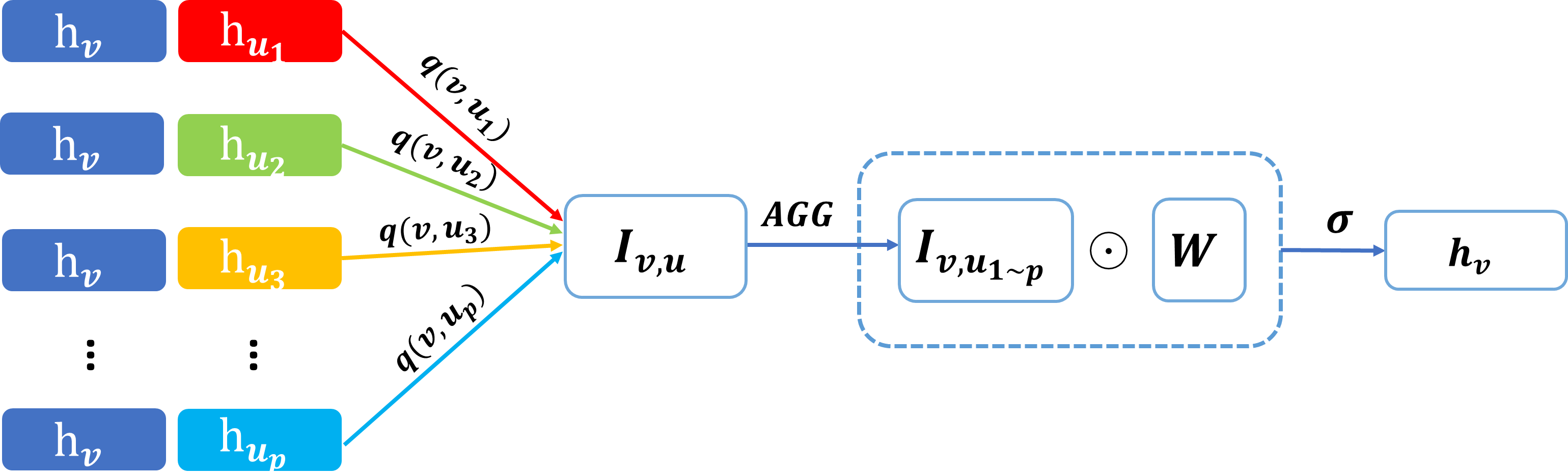}\\
		\caption{The position-aware embedding first aggregates the message of each target node $h_v$ and each anchor point $h_{u_i}$ in the anchor set via the aggregation function $I(v,u)$ based on the relative position coefficients $q_(v,u)$. The representation matrix is then further aggregated using the learnable function AGG and is finally transformed in a non-linear way to obtain the feature output $h_v$ for the target node in the current layer.}
		\label{figure3}
\end{figure}

Pixels at neighboring positions in the image convey similar semantic information, therefore we need to distinguish them to resolve matching ambiguities. Here, we refer to the nodes used as reference positions named as anchors and propose an effective strategy to construct an anchor set $P$ consisting of all nodes in the graph, denoted by $P=V$, serving as a stable reference for all nodes. To combine information from the nodes and the anchors, we design the information aggregation mechanism as is shown in Fig.\ref{figure3}.

Considering that the amount of effective information transmission to the nodes by the anchors with different relative distances is different, we first compute a relative position coefficient $q_{v,u}$ between a pair of nodes $v$ and $u$ as:

\begin{equation}\label{eq4}
q_{v,u}=\dfrac{e^{-d_{v,u}}}{\sum_{i \in V}{e^{-d_{v,i}}}},
\end{equation}
where $d_{v,u}$ is the shortest path distance between the two nodes.
In practice, to effectively reduce the time complexity of calculating the shortest path and also to reduce interference from those anchor
points that are distant from the node under consideration, we demand that the maximum shortest path distance does not exceed the ceiling value $r$. Otherwise, the value of the coefficient is set to infinity. We continue to compute  the information aggregation function $I(v,u)^{(l)}$ for the  $l$-th layer between node $v$ and $u$ as:

\begin{equation}\label{eq5}
I(v,u)^{(l)}=q_{v,u}{\rm{CONCAT}}(h_v^{(l-1)},h_u^{(l-1)}),
\end{equation}
where $h_v^{(l-1)}$ and $h_u^{(l-1)}$ are the  feature representation and position representation in the $(l-1)$-th layer, and  are combined through the message aggregation function. We further aggregate the information from all node-anchor pairs to obtaining a new representation in a high dimensional space using the non-linear variation. This hidden representation can be computed using the update 

\begin{equation}\label{eq6}
h_v^{(l)} = \sigma({\rm{AGG}}(I(v,u)^{(l)}| \ \forall u \in V) W^{(l)}),
\end{equation}
where AGG is typically a permutation-invariant function (e.g., sum), and $W^{(l)}$ is a learnable weight vector for the $l$-th layer. 


\subsection{Graph Feature Matching}\label{sec:4.3}
Utilizing the proposed position-aware node embedding, the proposed scheme encodes each node with position information into a high-level embedding space. 
In this way, we can simplify the second-order affinity matrix of paired graphs to a be a linear one with position-aware learning of the embedding. With the final hidden representation of the source graph $H_s \in \mathbb{R}^{n*F'}$ and the target graph $H_t \in \mathbb{R}^{m*F'}$ to hand, we can obtain a soft correspondence between these graphs with a node-wise affinity matrix through the inner product of the embeddings of the  two graphs being matched. Specifically, to satisfy the condition that the original graph maps injectively onto the target graph, we apply the Sinkhorn normalization \cite{sinkhorn1967concerning} to obtain rectangular doubly-stochastic correspondence matrices.

\begin{equation}\label{eq10}
S = \rm{sinkhorn}(H_sH_t^T).
\end{equation}

We further employ the cross-entropy loss function as the permutation loss between the predicted permutation matrix and the ground truth:

\begin{equation}\label{eq11}
\begin{aligned}
L = &-\sum_{i \in \mathcal{V}_{s}, j \in \mathcal{V}_{t}}\left(\mathbf{S}_{i, j}^{g t} \log \mathbf{S}_{i, j}+\left(1-\mathbf{S}_{i, j}^{g t}\right) \log \left(1-\mathbf{S}_{i, j}\right)\right),
\end{aligned}
\end{equation}
where $\mathbf{S}^{g t}$ is the ground truth permutation matrix. 
Consequently, the cross-entropy loss based on linear allocations can be learned end-to-end no matter the number of nodes and edges in the graph.

\begin{figure*}[t]
{\centering\includegraphics[width=1\textwidth]{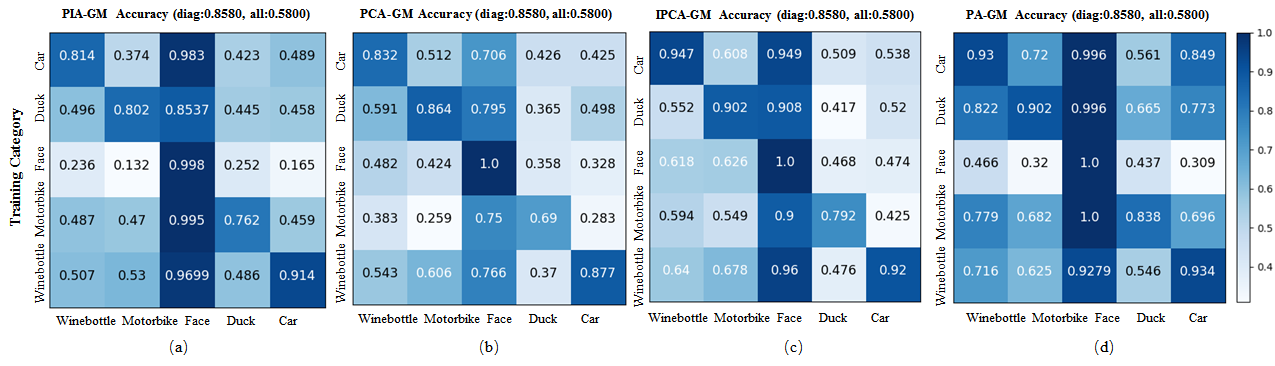}}
\caption{
Confusion matrix analysis of cross-category generalizations using the Willow ObjectClass dataset. The $y$-axis represents categories of images used for training the model, and the $x$-axis represents categories of samples used to test. Among them, (a), (b), and (c) are the comparison of the generalization ability of the aforementioned baseline in matching accuracy. (d) represents the method proposed in this paper.
}
\label{fig:Confusion Matrix1}
\vspace{-1.0em}
\end{figure*}

\section{Experiments}\label{sec:experiments}

\subsection{Experimental Setting}
\textbf{Dataset.} We use the Willow Object Class \cite{ChoICCV13}, consisting of 9963 images in total and is used as a benchmark by many methods to measure the accuracy of image classification and recognition algorithms. And IMC-PT-SparseGM \cite{JinIJCV21}  datasets, containing 16  object categories and 25061 images, which gather from 16 tourist attractions around the world. https://www.di.ens.fr/willow/research/graphlearning/.

\textbf{Implementation Details.} The experiments are conducted using two GeForce GTX 1080 Ti GPUs. We employ a batch size of 8 in training and evaluate the results of our method after 2000 epochs for each iteration. We employ the Adam \cite{kingma2014adam} optimizer to train our models with a learning rate of $1\times 10^{-4}$. To overcome the over-smoothing problem common to graph neural network models, we adopt a two-layer graph embedding and restrict the degree of smoothing in our experiments.

\textbf{Metrics.} We evaluate the graph matching capacity of the proposed method using the matching accuracy, which is defined  by $accuracy = \sum{{\rm{AND}}(S_{i,j},{S_{i,j}^{gt}})/N}$  from  \cite{WangICCV19,wang2020combinatorial}. Note that $S_{i,j}$ is the element of the predicted permutation matrix representing the correspondence matching of node $i$ and node $j$ from two different graphs. Similarly, $S_{i,j}^{gt}$ is the correspondence of the ground truth between two nodes, and $N$ is the number of matching node pairs.

\begin{table}[t]
	\centering
	\caption{Matching accuracy (\%) on the Willow ObjectClass.}
 	\renewcommand\arraystretch{1.2}
	\scalebox{0.8}
	{
	\begin{tabular}{c|c c c c c|c} 
		\hline
		Method&Car&Duck&Face&M-bike&W-bottle&Mean \\
		\hline
		GMN\cite{ZanfirCVPR18}&67.90&76.70&99.80&69.20&83.10&79.34\\
		
		NHGM\cite{WangPAMI21}&86.50&72.20&99.90&79.30&89.40&85.50\\
		
		CIE-H\cite{YuICLR20}&82.20&81.20&\textbf{100.00}&\textbf{90.00}&\textbf{97.60}&90.20\\
		
		PIA-GM\cite{WangICCV19}&88.60&87.00&\textbf{100.00}&70.30&87.80&86.74\\
		
		PCA-GM\cite{WangICCV19}&87.60&83.60&\textbf{100.00}&77.60&88.40&87.44\\
		
		IPCA-GM\cite{wang2020combinatorial}&90.40&88.60&\textbf{100.00}&83.00&88.30&90.06\\
		\hline
		\textbf{PA-GM (ours)}&\textbf{92.70}&\textbf{91.30}&\textbf{100.00}&84.50&93.80&\textbf{92.46}\\
		\hline

	\end{tabular}}
	\label{table2}
	\vspace{-1.0em}
\end{table}

\begin{table}[!]
	\centering
	\caption{Matching accuracy (\%) on the IMC-PT-SparseGM.}
 	\renewcommand\arraystretch{1.2}
	\scalebox{0.8}{
	\begin{tabular}{c|c c c|c} 
		\hline
		Method&Reichstag&Sacre\_coeur&St\_peters\_square&Mean\\
		\hline
		
		CIE-H\cite{YuICLR20}&42.24&28.47&30.78&33.83\\
		
		PIA-GM\cite{WangICCV19}&71.46&41.31&42.64&51.80\\
		
		PCA-GM\cite{WangICCV19}&69.38&39.86&42.10&50.40\\
		
		IPCA-GM\cite{wang2020combinatorial}&72.96&43.80&44.93&53.89\\
		
		GANN-GM\cite{wang2020graduated}&76.02&44.15&50.49&56.89\\
		\hline
		\textbf{PA-GM (ours)}&\textbf{96.28}&\textbf{75.93}&\textbf{81.66}&\textbf{84.63}\\
		\hline
	\end{tabular}}
	\label{table4}
	\vspace{-1.0em}
\end{table}

\subsection{Graph Matching Results on Real-world Datasets}
Table \ref{table2} and Table \ref{table4} report the overall comparison of the performance results for graph matching accuracy. The CIE-H method excels on the M-bike and W-bottle classes of the Willow ObjectClass dataset, while our method scores highest in the two categories of accuracy, and the average accuracy. By successfully incorporating the position information, our model achieves excellent results in the average accuracy of Willow ObjectClass and IMC-PT-SparseGM datasets. Based on these results, it can be concluded that our method performs well on graph matching compared with existing methods.

\subsection{Cross-category Generalization Study}\label{sec:generalization}
To assess the robustness and generalization ability of our method on the different object categories, we further conduct the cross-category generalization study. Specifically, we train each of the five classes separately, and then test the generalization ability of the validation model for all five classes with the separate training classes.  Comparing the elements of Fig. \ref{fig:Confusion Matrix1}, it is clear that the generalization ability of the model framework that incorporates position information is superior to the alternative methods. In addition, the matching accuracy for face data is generally rather large. This may be related to the relatively simple background of the data and less interference from noise.

\section{Conclusion}\label{sec:conclusion}
In this paper, we propose a novel deep learning method for graph matching based on node-wise embeddings between graphs that combine position-aware node embeddings. During the experiments, the proposed method is compared with alternative methods to demonstrate its robustness and effectiveness. Comparing our methodology to existing methods on real-world datasets demonstrates its state-of-the-art performance. Further improvements of graph matching will be achieved with the use of different relative positions strategies in future work.




\vfill\pagebreak



\bibliographystyle{IEEEbib}
\bibliography{strings,refs}

\end{document}